\documentclass[journal]{IEEEtran}

\usepackage[colorlinks,urlcolor=blue,linkcolor=blue,citecolor=blue]{hyperref}
\usepackage{amsmath,amssymb,amsfonts}
\usepackage{color,array}
\usepackage{graphicx}
\usepackage{multirow}
\usepackage{algorithm}
\usepackage{algorithmic}

\begin{document}

\title{VLMT: Vision-Language Multimodal Transformer for Multimodal Multi-hop Question Answering}

\author{Qi~Zhi~Lim, 
        Chin~Poo~Lee,~\IEEEmembership{Senior Member,~IEEE, }
        Kian~Ming~Lim,~\IEEEmembership{Senior Member,~IEEE, }
        Kalaiarasi~Sonai~Muthu~Anbananthen
\thanks{Qi Zhi Lim is with the Faculty of Information Science and Technology, Multimedia University, Jalan Ayer Keroh Lama, 75450 Melaka, Malaysia (e-mail: 1181103589@student.mmu.edu.my).}
\thanks{Chin Poo Lee is with the School of Computer Science, University of Nottingham Ningbo China, 199 Taikang East Road, Yinzhou District, Ningbo, Zhejiang Province, 315100, China (e-mail: leechinpoo@outlook.com). Corresponding author: Chin Poo Lee.}
\thanks{Kian Ming Lim is with the School of Computer Science, University of Nottingham Ningbo China, 199 Taikang East Road, Yinzhou District, Ningbo, Zhejiang Province, 315100, China (e-mail: Kian-Ming.Lim@nottingham.edu.cn).}
\thanks{Kalaiarasi Sonai Muthu Anbananthen is with the Faculty of Information Science and Technology, Multimedia University, Jalan Ayer Keroh Lama, 75450 Melaka, Malaysia (e-mail: kalaiarasi@mmu.edu.my).}}

\markboth{IEEE Transactions on [Journal Name],~Vol.~00, No.~0, [Month]~2020}%
{Lim \MakeLowercase{\textit{et al.}}: VLMT: Vision-Language Multimodal Transformer for Multimodal Multi-hop Question Answering}

\maketitle

\begin{abstract}
The increasing availability of multimodal data across text, tables, and images presents new challenges for developing models capable of complex cross-modal reasoning. Existing methods for Multimodal Multi-hop Question Answering (MMQA) often suffer from limited reasoning capabilities, reliance on modality conversion (e.g., image-to-text), and inadequate alignment between visual and textual representations. To address these limitations, this paper introduces Vision-Language Multimodal Transformer (VLMT), a unified architecture that integrates a transformer-based vision encoder with a sequence-to-sequence language model. VLMT employs a direct token-level injection mechanism to fuse visual and textual inputs within a shared embedding space, eliminating the need for intermediate projection layers. To enhance cross-modal alignment and reasoning, a three-stage pretraining strategy is proposed to progressively align vision-language representations and improve the model’s capacity for multimodal understanding. Based on the pretrained backbone, two task-specific modules are instantiated to form a two-stage MMQA framework: a multimodal reranker that predicts document relevance scores and utilizes a relative threshold with top-$k$ strategy for context retrieval, and a multimodal question answering model that generates contextually grounded answers based on the retrieved evidence. Comprehensive experiments on two benchmark datasets demonstrate the effectiveness of the proposed approach. On MultimodalQA validation set, VLMT-Large achieves 76.5\% Exact Match and 80.1\% F1, outperforming the previous state-of-the-art by +9.1\% in Exact Match and +8.8\% in F1. On WebQA, it attains a QA score of 47.6, surpassing prior models such as PERQA by +3.2. These results highlight VLMT’s strong capabilities in multimodal reasoning and its potential to advance real-world information retrieval and question answering systems.
\end{abstract}


\section{Introduction}
\label{sec:introduction}

The exponential growth of information in today’s digital ecosystem has led to the proliferation of multimodal data—comprising text, tables, and images—across a wide range of platforms. This surge in heterogeneous content offers unprecedented opportunities for knowledge extraction, but also introduces challenges for tasks that require joint understanding and reasoning across diverse modalities.

Multimodal Multi-hop Question Answering (MMQA)~\cite{talmor2021mmqa, chang2022webqa} has emerged as a representative task in this domain, reflecting real-world information-seeking behavior where relevant evidence is scattered across multiple sources and modalities. MMQA requires models to perform two interdependent operations: retrieving relevant multimodal context and reasoning over the retrieved information to produce accurate and coherent answers. The dual nature of MMQA—retrieval and reasoning—necessitates robust cross-modal integration and effective multi-hop inference.

Early solutions to MMQA have largely followed modular paradigms. Some approaches employ modality-specific models by classifying question types or decomposing complex queries into simpler sub-questions~\cite{talmor2021mmqa, rajabzadeh2023multimodal}. While effective in unimodal settings, these strategies often suffer from insufficient cross-modal interaction and fail to capture interdependencies among modalities. Other works address this limitation by converting non-textual content—particularly images—into textual descriptions through captioning or object detection~\cite{yu2023unified, lim2024unirag}. This enables the use of pre-trained language models for both retrieval and answer generation. However, such conversion pipelines introduce a dependency on transformation quality and may incur semantic loss, impairing the model’s ability to capture fine-grained visual details.

Recent advances in multimodal learning have led to the development of models capable of processing multiple modalities simultaneously~\cite{chen2022murag, yang2023enhancing, yang2023progressive}. These models integrate visual and textual features within unified architectures to enable joint reasoning. Nevertheless, robust alignment between modalities remains a persistent challenge, often resulting in degraded performance on tasks that require precise semantic grounding and fine-grained evidence aggregation.

To overcome these limitations, this paper introduces \textit{Vision-Language Multimodal Transformer} (VLMT), a unified multimodal architecture designed specifically for MMQA. VLMT combines a transformer-based vision encoder with a sequence-to-sequence language model. It leverages a direct token-level injection mechanism, wherein visual embeddings are inserted into designated positions in the textual input. This design enables seamless multimodal fusion without the need for intermediary projection layers or modality-specific adapters, thereby preserving efficiency and reducing complexity.

Given the architectural heterogeneity between the vision and language components, effective cross-modal integration requires dedicated pretraining. To this end, a three-stage pretraining framework is proposed to progressively enhance multimodal capabilities. The first stage aligns visual embeddings with the frozen embedding space of the language model using instruction-following image-caption tasks. The second stage performs joint optimization using semantically enriched image-text pairs to refine fine-grained alignment. The final stage focuses on visual question answering, during which the language model is trained to perform multimodal reasoning while the vision encoder remains fixed. This progressive alignment strategy enhances the model’s ability to integrate and reason over multimodal information, ultimately improving its performance in downstream tasks.

Building on the pretrained VLMT backbone, a two-stage MMQA framework is constructed. The first stage employs a multimodal reranker that scores candidate documents based on their relevance to a given question, and applies top-$k$ strategy in conjunction with relative threshold to retrieve informative contexts. The second stage uses a multimodal question answering model that generates answer based on the selected content. Both modules operate in a fully multimodal fashion, sharing the VLMT backbone and benefiting from its robust alignment and reasoning capabilities.

Extensive experiments on the MultimodalQA and WebQA datasets demonstrate that the proposed framework achieves state-of-the-art performance, significantly outperforming existing methods in both retrieval and answer generation. These results validate the effectiveness of the proposed architecture and training strategy, establishing VLMT as a robust solution for complex multimodal reasoning tasks. The main contributions of this work are summarized as follows:

\begin{itemize}
    \item \textbf{Unified Multimodal Architecture:} A novel Vision-Language Multimodal Transformer (VLMT) is proposed, which integrates a transformer-based vision encoder and a sequence-to-sequence language model within a unified architecture. VLMT employs a direct token-level injection mechanism to fuse visual and textual inputs in a shared embedding space, eliminating the need for intermediate projection modules.

    \item \textbf{Progressive Pretraining Framework:} A dedicated three-stage pretraining strategy is introduced to progressively align vision and language representations. This framework strengthens multimodal integration and equips the model with strong visual-semantic reasoning capabilities critical for MMQA tasks.

    \item \textbf{Task-Specific MMQA Framework:} A two-stage MMQA framework is developed, consisting of a multimodal reranker for context retrieval and a multimodal question answering model for answer generation. Both components are instantiated from the pretrained VLMT backbone and adapted to support document ranking and question answering, respectively.

    \item \textbf{Scalable Design and Empirical Validation:} Two VLMT configurations—VLMT-Base and VLMT-Large—are introduced to support varying computational budgets. Extensive experiments on the MultimodalQA and WebQA benchmarks demonstrate that both configurations outperform existing methods, with VLMT-Large achieving new state-of-the-art results in both datasets.
\end{itemize}

The remainder of this paper is organized as follows. Section~\ref{sec:related_work} provides an overview of related work in multimodal multi-hop question answering and multimodal large language models. Section~\ref{sec:methodology} introduces the proposed VLMT framework, detailing its unified architecture, progressive pretraining strategy, and task-specific components for reranking and question answering. Section~\ref{sec:experiments} outlines the experimental setup, including datasets, implementation details, inference procedures, and ablation studies, followed by a comprehensive evaluation on benchmark datasets. Finally, Section~\ref{sec5:conclusion} concludes the paper and outlines promising directions for future work.

\section{Related Work}
\label{sec:related_work}

This section examines research efforts most relevant to this study, focusing primarily on two key areas: multimodal multi-hop question answering (MMQA) and multimodal large language models (MLLMs). MMQA extends the scope of traditional question-answering by requiring multi-step reasoning across different modalities, necessitating both advanced retrieval mechanisms and effective answer generation strategies. In parallel, MLLMs leverage large-scale vision-language modeling techniques to facilitate deep multimodal understanding and cross-modal reasoning. The following subsection surveys the key advancements and methodologies in these areas, highlighting their technical contributions and associated limitations in addressing MMQA challenges.

\subsection{Multimodal Multi-hop Question Answering}

The task of multimodal question answering has attracted increasing attention due to its requirement to integrate and reason over information from distinct modalities. This task generalizes earlier text-only question answering paradigms~\cite{rajpurkar2016squad} to include visual and structured sources, thereby demanding more comprehensive understanding and reasoning capabilities. The Visual Question Answering (VQA) dataset~\cite{antol2015vqa} served as an early benchmark in this field by pairing natural language questions with images. Follow-up datasets such as OK-VQA~\cite{marino2019ok} and KVQA~\cite{shah2019kvqa} further challenged models by requiring external knowledge in addition to visual grounding to generate accurate answers. Other studies explored reasoning across tables and text, as seen in HybridQA~\cite{chen2020hybridqa} and TAT-QA~\cite{zhu2021tat}, highlighting the importance of modeling interactions across structured and unstructured modalities.

Building on these efforts, recent works have introduced the MMQA task~\cite{talmor2021mmqa, chang2022webqa, hannan2020manymodalqa}, which poses additional complexity by requiring reasoning over multiple steps and modalities simultaneously. Among existing datasets, MultimodalQA~\cite{talmor2021mmqa} and WebQA~\cite{chang2022webqa} are the most widely adopted resources for MMQA research. Unlike conventional settings in which the relevant context is provided exclusively, these datasets operate under open-domain conditions. Thus, models must first retrieve relevant content from disparate sources and subsequently perform answer generation, often involving multi-hop reasoning across diverse modalities.

In response to the intricate requirements of MMQA, several baseline methods were introduced. AutoRouting and ImplicitDecomp~\cite{talmor2021mmqa} are two such models designed for MultimodalQA. In AutoRouting, a question-type classifier is used to determine the target modality, and the question along with candidate contexts is routed to the appropriate single-modality question answering (QA) module. ImplicitDecomp employs a two-hop reasoning process in which a classifier predicts the question type, including the required modalities, reasoning order, and operations. At each hop, a combination of the original question, hop count, modality-specific contexts, and intermediate answers is passed into the corresponding QA module for answer generation. In both cases, the text and table QA modules are based on RoBERTa-large~\cite{liu2019roberta}, while the image QA module utilizes VILBERT-MT~\cite{lu202012} with Faster R-CNN~\cite{ren2016faster} for visual feature extraction.

For WebQA dataset, VLP and VLP+VinVL~\cite{chang2022webqa} have been proposed as baseline models. These generative architectures are based on encoder-decoder transformers, initialized from the pre-trained VLP backbone~\cite{zhou2020unified}. VLP+VinVL extends this architecture by integrating improved visual representations from VinVL~\cite{zhang2021vinvl}. The overall system is divided into two specialized components: a source retrieval module and a question answering module. In the retrieval phase, each source is concatenated with the question and scored using a VLP-based classifier. The top-ranked sources are then passed into the QA model for answer generation.

Several more recent approaches have been developed to improve multimodal integration and reasoning. MuRAG~\cite{chen2022murag} constructs an external memory using T5-base word embeddings~\cite{raffel2020exploring} and ViT-large~\cite{dosovitskiy2020image} visual representations. Maximum inner product search is utilized for context retrieval, and the T5 encoder-decoder is responsible for answer generation. 

Another example is SKURG~\cite{yang2023enhancing}, which encodes multimodal inputs into a unified semantic space and incorporates an entity-centric fusion mechanism. The architecture uses OFA-base~\cite{wang2022ofa} as the vision encoder and BART-base~\cite{lewis2019bart} for both text and table encoding. Named entity recognition~\cite{peters2017semi} and relation extraction~\cite{han2019opennre} are employed to derive structured knowledge for enhanced performance.

In contrast, Solar~\cite{yu2023unified} proposes a unified language-space framework that converts tables into templated sentences and transforms images into text using BLIP~\cite{li2022blip} and VinVL~\cite{zhang2021vinvl}. The model follows a three-stage process involving retrieval, ranking, and generation. BERT model~\cite{devlin2018bert} is employed to support the retrieval and ranking stages, while the answer generation stage is handled using T5 model~\cite{raffel2020exploring}.

PERQA~\cite{yang2023progressive} adopts a progressive evidence refinement strategy composed of an initial screening module and an iterative retrieval mechanism. Visual descriptions are extracted using OFA-large~\cite{wang2022ofa} and Fast-RCNN~\cite{girshick2015fast}, while BART-base~\cite{lewis2019bart} and DeBERTa-large~\cite{he2020deberta} are used for encoding and screening evidence. The multi-turn QA stage utilizes mPlug-Owl~\cite{ye2023mplug}, a vision-language model built upon ViT and LLaMA~\cite{touvron2023llama}, fine-tuned with low-rank adaptation~\cite{hu2021lora}.

UniRaG~\cite{lim2024unirag} introduces a three-stage pipeline comprising unification, retrieval, and generation. The model converts all modalities into text, employing LLaVA~\cite{liu2024improved} for image captioning to minimize information loss. A BERT-based classifier (ms-marco-MiniLM-L-12-v2~\cite{reimers2019sentence}) is fine-tuned for retrieval, while Flan-T5-Base~\cite{chung2024scaling} is used for answer generation.

Despite substantial progress, existing models often require complex pre-processing pipelines or rely heavily on modality transformation quality. These dependencies may result in information degradation and introduce challenges for real-world deployment. Limitations also persist in terms of efficient alignment and reasoning across modalities, suggesting that further research is needed to improve both the performance and practicality of MMQA systems.

\subsection{Multimodal Large Language Models}

Large language models (LLMs) have demonstrated exceptional capabilities in understanding and generating human language, significantly advancing the field of natural language processing across diverse applications. Building on these developments, vision-language models have been introduced to extend the utility of LLMs by incorporating visual inputs. These models enable multimodal understanding and support a wide range of tasks such as image captioning, visual reasoning, and optical character recognition.

Recent progress has led to the emergence of multimodal large language models (MLLMs)~\cite{liu2024visual, liu2024improved}, which have become a prominent paradigm for addressing multimodal tasks. MLLMs typically integrate a pre-trained language model with a vision encoder within a unified architecture. Instruction-following capability is considered essential for handling various downstream tasks, and therefore instruction-tuned LLMs~\cite{chung2024scaling, zheng2023judging} are frequently adopted as the language backbone in these models. In parallel, visual encoders~\cite{radford2021learning, zhai2023sigmoid} are utilized to extract semantically rich representations from visual data. To enable seamless interaction between visual and textual modalities, a projector module is commonly used to map visual features into the language embedding space. This configuration allows the language model to jointly process and reason over multimodal inputs.

While MLLMs exhibit strong generalization capabilities and have achieved state-of-the-art performance on several vision-language benchmarks, their application to domain-specific tasks such as multimodal multi-hop question answering (MMQA)~\cite{talmor2021mmqa, chang2022webqa} remains limited. MMQA introduces distinct challenges, including open-domain retrieval from heterogeneous modalities and multi-step reasoning over diverse information sources. Existing MLLM architectures are often not optimized for fine-grained modality alignment or scalable context retrieval, which are critical for accurate and contextually grounded answer generation in MMQA.

\begin{figure*}[htbp]
    \centering
    \includegraphics[width=\textwidth]{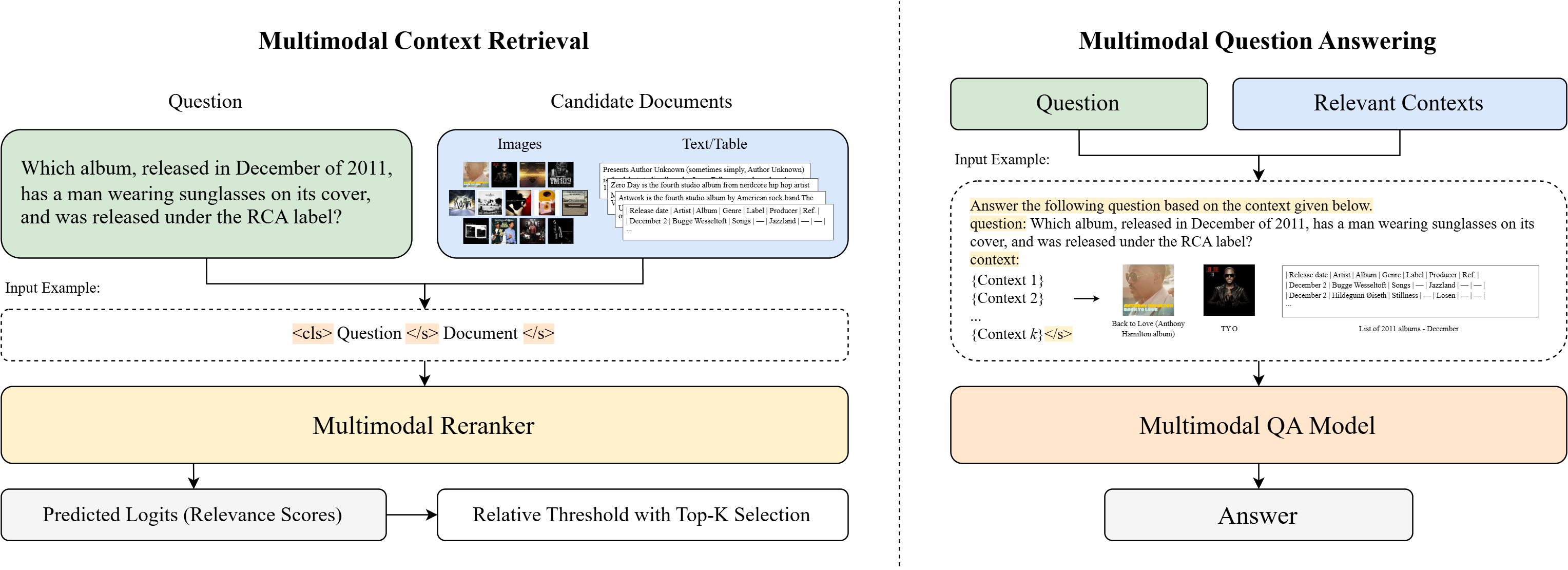}
    \caption{The proposed two-stage framework for MMQA. The first stage (left) employs a multimodal reranker to evaluate candidate documents and selects relevant contexts using a relative threshold with a top-$k$ strategy. The second stage (right) utilizes a multimodal QA model to generate contextually coherent answers based on the input question and the retrieved multimodal contexts.}
    \label{fig:proposed_framework}
\end{figure*}

Despite the progress in MLLMs, there remains a lack of unified frameworks specifically tailored for MMQA tasks. Current models either rely heavily on pre-processing pipelines or depend on modality-specific transformations, which may lead to information loss and suboptimal cross-modal reasoning. Moreover, most MLLMs are not pretrained with objectives explicitly designed for visual reasoning in multi-hop settings. These limitations underscore the need for a specialized multimodal model that supports direct integration of vision and language features, incorporates progressive pretraining for vision-language alignment, and enables efficient retrieval and reasoning in MMQA scenarios. The proposed VLMT framework is introduced to address this gap.

\section{Methodology}
\label{sec:methodology}

This section presents the proposed Vision-Language Multimodal Transformer (VLMT), a unified multimodal backbone model designed to address the challenges of MMQA. VLMT comprises a vision encoder and an instruction-tuned language model within a cohesive framework, enabling direct interaction between modalities through a token-level injection mechanism that eliminates the need for intermediate projection or transformation layers. To further enhance vision-language alignment and reasoning capabilities, a three-stage pretraining framework is introduced, equipping VLMT with strong multimodal representations suitable for complex downstream tasks.

Building on the pretrained VLMT backbone, a task-specific two-stage framework is constructed to handle MMQA. As illustrated in Fig.~\ref{fig:proposed_framework}, the first stage employs a multimodal reranker to rank candidate documents based on their relevance to a given question. A relative threshold strategy with a top-$k$ filter is applied to retrieve semantically meaningful contexts. The second stage utilizes a multimodal QA model that processes the input question along with the retrieved contexts to generate accurate and contextually grounded answers.

Unlike most conventional MMQA pipelines that convert images into textual descriptions, the proposed framework enables implicit multimodal reasoning by directly processing images, text, and tables without explicit modality conversion. Both the reranker and QA model are instantiated from the pretrained VLMT backbone through task-specific adaptations. The remainder of this section details the core architecture of VLMT, the pretraining strategy, and the implementation of the reranker and QA modules.

\subsection{Vision-Language Multimodal Transformer (VLMT)}
\label{sec:vlmt_backbone}

The Vision-Language Multimodal Transformer (VLMT) serves as the backbone architecture of the proposed framework, designed to enable unified representation learning and cross-modal reasoning over both visual and textual modalities. VLMT is composed of a transformer-based vision encoder for extracting spatial and semantic representations from images, and a sequence-to-sequence language model that supports contextual understanding and natural language generation. These two components are jointly structured within a shared embedding space to facilitate seamless multimodal integration without the need for intermediary projection layers.

Unlike conventional multimodal models that require image-to-text transformations or learned projection modules to connect heterogeneous modalities, VLMT achieves alignment by ensuring that the vision and language components operate on embeddings of equal dimensionality. This design allows direct fusion of visual and textual information through a token-level injection mechanism, which introduces visual embeddings into the token sequence at predefined positions. The resulting representation is processed by the language model’s attention mechanism, enabling cross-modal interaction without additional architectural complexity.

\begin{figure}[htbp]
    \centering
    \includegraphics[width=\columnwidth]{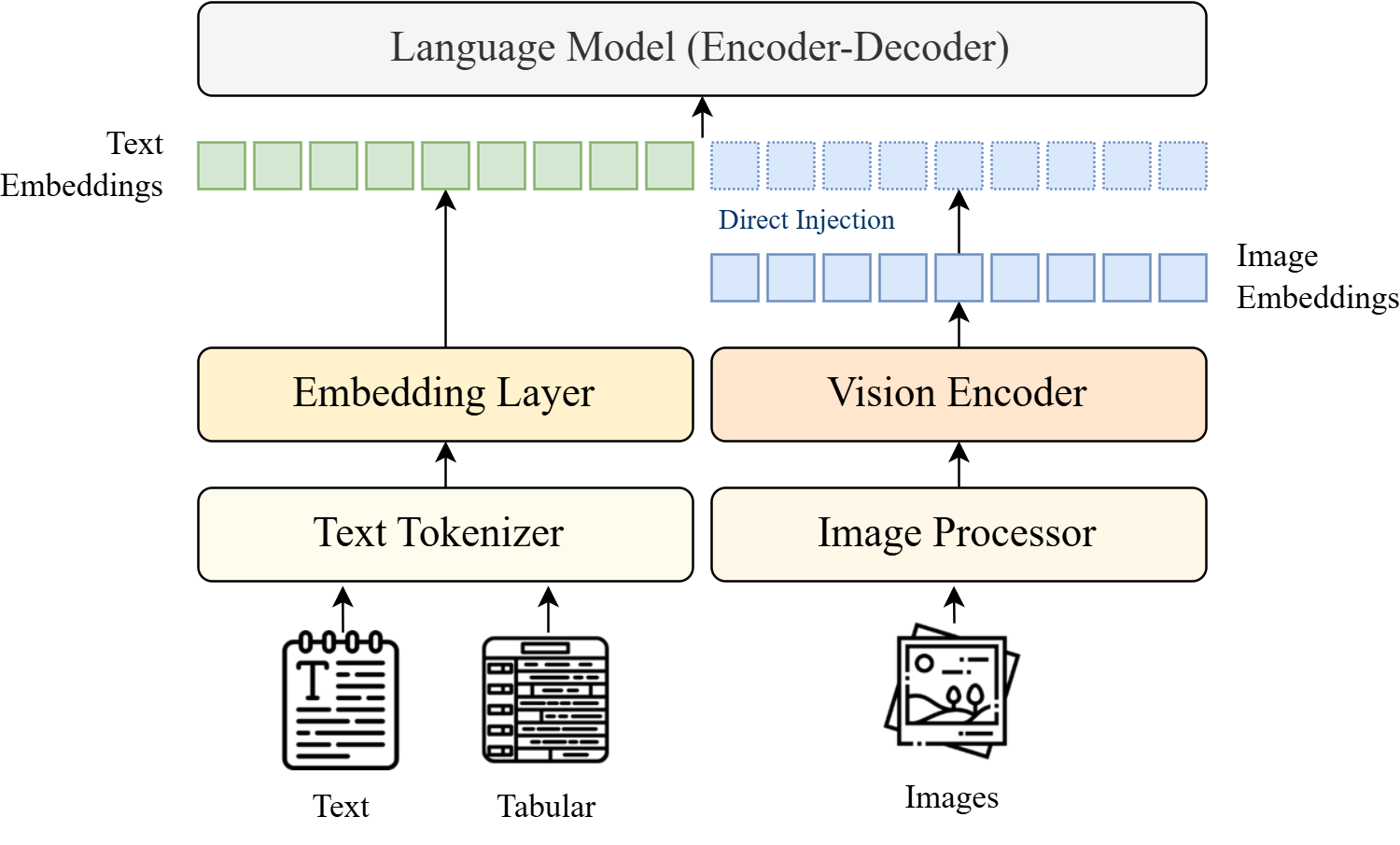}
    \caption{Architecture of VLMT. Textual inputs are processed via tokenization and embedding layers, while visual inputs are passed through a vision encoder. Given a shared embedding dimension, image embeddings are directly injected into the input token sequence by replacing designated visual placeholders, enabling seamless and efficient multimodal fusion.}
    \label{fig:vlmt_architecture}
\end{figure}

The overall structure of VLMT is illustrated in Fig.~\ref{fig:vlmt_architecture}. Text inputs are tokenized and mapped to embeddings via a subword-based encoder, while image inputs are preprocessed and encoded into fixed-length visual tokens. These image embeddings are inserted into the text embedding sequence by replacing designated visual placeholder tokens. This mechanism supports unified encoding of multimodal content and allows joint reasoning over both modalities.

\subsubsection{Multimodal Data Representation and Integration}
\label{sec:vlmt_mdri}

The encoding process in VLMT begins by independently processing visual and textual inputs through dedicated pipelines. For textual data, including plain text and serialized tables, inputs are tokenized using a subword-level model and mapped to dense vector representations through an embedding layer that capture both semantic and structural information.

\begin{figure}[htbp]
    \centering
    \includegraphics[width=\columnwidth]{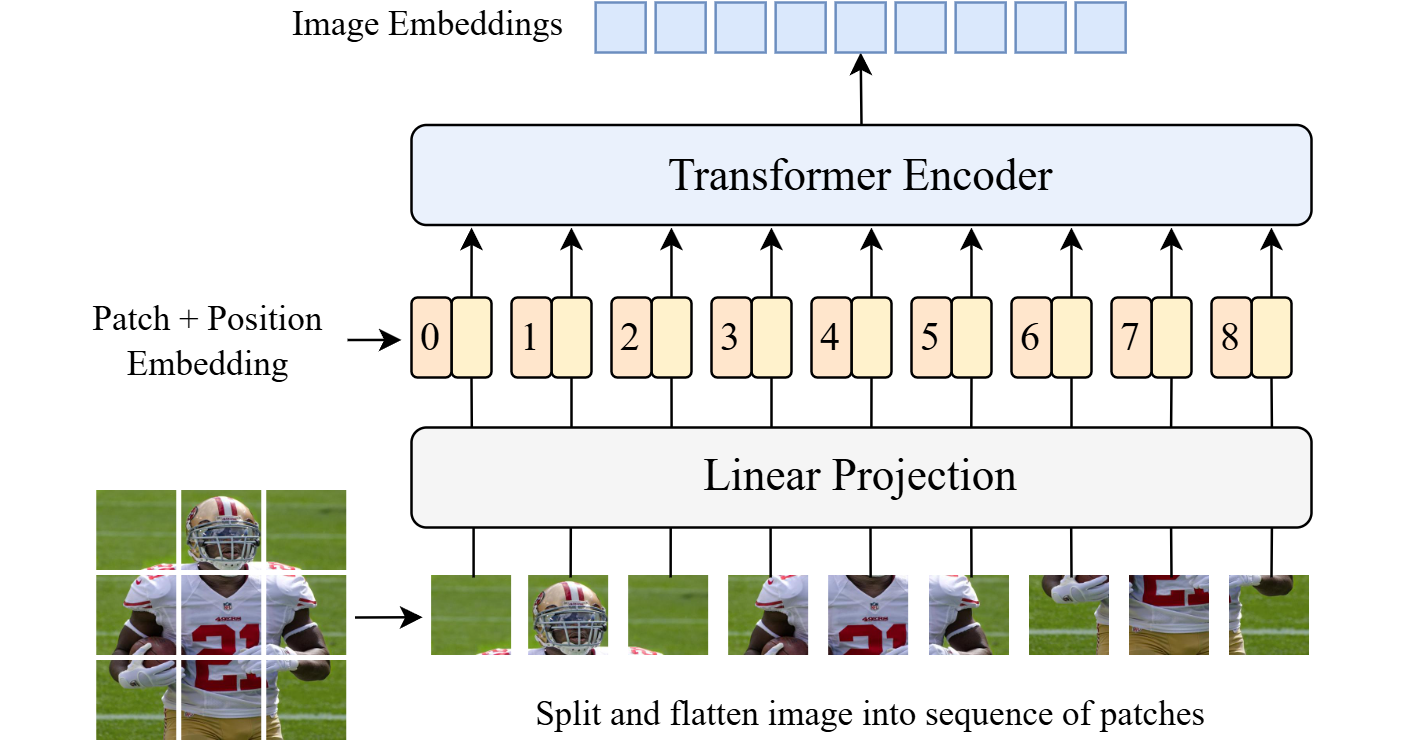}
    \caption{Visual feature extraction using the vision encoder. Images are divided into patches, embedded, and encoded via transformer layers to produce spatially aware representations.}
    \label{fig:vision_encoder}
\end{figure}

Fig.~\ref{fig:vision_encoder} illustrates the visual feature extraction pipeline. Visual inputs are processed using a transformer-based architecture that segments image into a grid of uniform patches. Let $x \in \mathbb{R}^{H \times W \times C}$ denote the input image, which is divided into $N$ non-overlapping patches of size $P \times P$, resulting in a sequence $x_p \in \mathbb{R}^{N \times (P^2 \cdot C)}$. Each patch is flattened and linearly projected into the shared embedding space. Additionally, positional embeddings are added to maintain the spatial relationships among patches. The resulting sequence of embeddings is then passed through a stack of transformer layers to generate the image embeddings.

After generating both textual and visual embeddings, multimodal integration is performed by inserting the image embeddings into the text embedding sequence at predefined positions. Formally, $\mathbf{E}_{\text{text}} \in \mathbb{R}^{L \times d}$ denote the text embeddings, and $\mathbf{E}_{\text{image}} \in \mathbb{R}^{N \times d}$ denote the image embeddings. Given the designated positions for image tokens $\{i_1, i_2, \dots, i_N\}$, the fused representation $\mathbf{E}_{\text{fused}}$ is defined as:

\begin{equation}
\label{eq:direct_injection}
\begin{gathered}
\mathbf{E}_{\text{fused}}[i_j, :] = \mathbf{E}_{\text{image}}[j, :], \quad \forall j \in [1, N], \\
\mathbf{E}_{\text{fused}}[k, :] = \mathbf{E}_{\text{text}}[k, :], \quad \forall k \notin \{i_1, i_2, \ldots, i_N\}.
\end{gathered}
\end{equation}

This direct injection strategy eliminates the need for intermediate fusion layers or modality-specific adapters, enabling efficient and scalable multimodal integration. The fused embedding sequence is then passed to the language model component of VLMT, which employs encoder-decoder architecture with cross-modal attention mechanisms to  facilitate contextual reasoning and enable downstream generation.

This architecture supports end-to-end training and inference over heterogeneous inputs while preserving the fine-grained semantics and structural integrity of each modality. It forms the foundation for both retrieval and generation tasks within the broader multimodal multi-hop question answering framework.

\subsubsection{VLMT Pretraining Framework}
\label{sec:vlmt_pretraining_framework}

To support robust multimodal reasoning and effective vision-language integration, a three-stage pretraining framework is introduced for VLMT. The first two stages are designed to progressively align visual and linguistic representations, addressing the inherent inconsistency that arises when combining independently trained vision and language components. The final stage focuses on enhancing the model’s capacity for visual question answering (VQA), enabling more accurate and contextually grounded reasoning over multimodal inputs.

In the first stage, the model is trained using the LLaVA Visual Instruct Pretrain LCS-558K dataset~\cite{liu2024improved}, which reformulates image-caption pairs into instruction-following tasks. Each training sample consists of an image paired with a textual prompt requesting a brief description, while the original caption serves as the target output. During this phase, only the vision encoder is updated, while the language model remains frozen.
This selective training encourages the vision encoder to produce embeddings compatible with the language model’s pre-trained word representations. However, the limited descriptiveness of the captions may constrain the depth of visual-linguistic alignment achieved.

To overcome this limitation, the second stage employs ShareGPT4V-PT~\cite{chen2025sharegpt4v}, a high-quality dataset comprising 1.2 million image-caption pairs. The captions include detailed semantic content of the images, such as object attributes, spatial relations, and world knowledge. In this phase, both the vision encoder and language model are jointly optimized, promoting fine-grained alignment across modalities. The vision encoder learns to capture richer visual features, while the language model learns to interpret and integrate this information, resulting in a more coherent and semantically aligned multimodal representation space.

The third stage focuses on enhancing the model's reasoning capabilities through VQA pretraining using the VQA v2.0 dataset~\cite{goyal2017making}. This dataset contains open-ended questions that require both visual understanding and contextual reasoning. To retain the visual features learned in earlier stages, the vision encoder is kept frozen, and only the language model is fine-tuned in this stage. This targeted adaptation enables the model to reason over visual inputs conditioned on natural language queries, improving its ability to generate accurate and contextually grounded answers.

This three-stage pretraining framework equips VLMT with strong cross-modal alignment and visual reasoning capabilities, forming a robust foundation for downstream tasks such as multimodal multi-hop question answering.

\subsection{VLMT as Multimodal Reranker}
\label{sec:vlmt_reranker}

The first stage of the proposed MMQA framework incorporates a multimodal reranker to identify supporting documents from a candidate pool. Accurate identification of relevant context is essential for minimizing computational overhead during answer generation and enhancing the overall reasoning accuracy. To this end, the reranker computes relevance scores for each candidate document and applies a novel selection strategy that combines a relative threshold with a top-$k$ constraint. This dual mechanism balances the retrieval precision and recall by dynamically adapting to the score distribution while ensuring selection focus and diversity.

Given a question and a candidate document, the input sequence is constructed by concatenating them with task-specific control tokens. A \texttt{<cls>} token is prepended to serve as a global sequence representation, and \texttt{</s>} tokens are inserted to separate question and document segments and denote the sequence boundaries.

The reranker leverages the encoder of the pretrained VLMT backbone to process the input sequence. Since the reranking task involves classification rather than generation, only the encoder component is utilized. After encoding, the hidden state corresponding to the \texttt{<cls>} token is extracted and denoted as $h \in \mathbb{R}^{d_{\text{enc}}}$, where $d_{\text{enc}}$ is the embedding dimension of the encoder. This token serves as a global representation summarizing the entire input sequence.

To compute a scalar relevance score from $h$, a lightweight classification head operates in three stages is applied. The scoring function is defined as follows:

\begin{equation}
\label{eq:cls_head}
y = W_2 \cdot \operatorname{Dropout} \left( \tanh \left( W_1 \cdot \operatorname{Dropout}(h) + b_1 \right) \right) + b_2,
\end{equation}
where $W_1 \in \mathbb{R}^{d \times d}$ and $b_1 \in \mathbb{R}^{d}$ are the weights and bias of the intermediate linear transformation, and $W_2 \in \mathbb{R}^{1 \times d}$ and $b_2 \in \mathbb{R}$ are the parameters of the final output layer. 

 First, a dropout operation is applied to $h$ for regularization. Then, the transformed vector $W_1 \cdot h + b_1$ is passed through a $\tanh$ activation to introduce non-linearity. A second dropout layer is applied to the output of the activation, followed by a final linear projection to obtain the output logit $y \in \mathbb{R}$. This scalar score reflects the model's estimation of the document's relevance to the input question.

The raw logits are passed through a sigmoid function to normalize them to the range $[0, 1]$. Let $\hat{y}_i$ denote the normalized relevance score for the $i$-th candidate in a given set of $M$ candidates. Let $\hat{y}_{\max} = \max_{i \in [1, M]} \hat{y}_i$ be the maximum relevance score in the candidate set. The proposed relative thresholding strategy selects documents that satisfy:

\begin{equation}
\label{eq:rel_thresh}
\hat{y}_i \geq \tau \cdot \hat{y}_{\max}, \quad \forall i \in [1, M],
\end{equation}
where $\tau \in (0, 1]$ is a predefined threshold ratio. This criterion ensures that only candidates achieving a substantial proportion of the highest score are considered, making the selection adaptive to score distributions across different queries.

To avoid an excessively large context set, the candidate pool is further refined using a top-$k$ selection, which retains at most $k$ documents from those satisfying Eq.~\eqref{eq:rel_thresh}. The final context set $\mathcal{D}_{\text{retrieved}}$ is defined as:

\begin{equation}
\label{eq:topk_select}
\mathcal{D}_{\text{retrieved}} = \text{Top-}k\left(\left\{ d_i \mid \hat{y}_i \geq \tau \cdot \hat{y}_{\max} \right\}\right).
\end{equation}

This two-stage filtering approach is central to the reranker’s design. The relative threshold offers adaptivity across questions with varying score distributions, while top-$k$ ensures bounded retrieval cost. Together, they form a flexible and efficient strategy for multimodal context selection.

\subsubsection{Training Objective}

The reranker is trained as a binary classifier. Each question is paired with a batch of $N$ documents, including one or more positive (supporting) samples labeled as $y_i = 1$, and several distractors labeled as $y_i = 0$. The objective is to distinguish the true supporting evidence from irrelevant content, which is optimized using the binary cross-entropy with logits loss:

\begin{equation}
\label{eq:cls_loss}
\mathcal{L}_{\text{cls}} = -\frac{1}{N} \sum_{i=1}^N \left[ y_i \cdot \log \left( \sigma(\hat{y}_i) \right) + (1 - y_i) \cdot \log \left( 1 - \sigma(\hat{y}_i) \right) \right],
\end{equation}
where $\sigma$ denotes the sigmoid function applied to the predicted logit $\hat{y}_i$. This loss formulation guides the model to maximize scores for supporting documents while minimizing scores for distractors, thereby improving the reranker's precision and recall in context retrieval.

\subsection{VLMT as Multimodal Question Answering Model}
\label{sec:vlmt_qa_model}

The second stage of the proposed MMQA framework employs a multimodal QA model to generate responses based on the retrieved multimodal contexts. Given the inherently multimodal and multi-hop characteristics of the task, the model input is formulated by prepending the question to a sequence of retrieved documents, which may include text, table, and image-derived content. Instructional prompts are prepended to guide the model toward generating contextually relevant and semantically coherent answers, as depicted in Fig.~\ref{fig:proposed_framework}.

The QA model is instantiated from the full encoder-decoder architecture of the pretrained VLMT backbone. Unlike the reranker, which requires only discriminative capabilities, this stage necessitates generative modeling; thus, both the encoder and decoder components are utilized. The encoder processes the input sequence to produce contextualized hidden states, while the decoder performs autoregressive generation of the answer by attending to these representations.

Let $X = \{x_1, x_2, \ldots, x_L\}$ represent the input token sequence, comprising the question and associated multimodal context. The encoder maps $X$ to a sequence of contextualized embeddings $\mathbf{H} = \{\mathbf{h}_1, \mathbf{h}_2, \ldots, \mathbf{h}_L\}$, which are attended to by the decoder at each generation step.

A language modeling head is appended to the decoder, projecting each decoder hidden state $\mathbf{z}_t \in \mathbb{R}^{d_{\text{dec}}}$ into the tokenizer’s vocabulary space. Specifically, the probability of generating the $t$-th token $y_t$ is defined as:
\begin{equation}
P(y_t \mid X, y_{<t}) = \operatorname{Softmax}(W_o \cdot \mathbf{z}_t + b_o),
\end{equation}
where $W_o \in \mathbb{R}^{V \times d_{\text{dec}}}$ and $b_o \in \mathbb{R}^{V}$ are the output projection weights and bias, $V$ denotes the vocabulary size, and $d_{\text{dec}}$ is the dimensionality of the decoder hidden state.

\subsubsection{Training Objective}

The training of the multimodal QA model is formulated as a sequence-to-sequence generation task. For each instance, the input consists of the question and its corresponding multimodal context, and the output is the ground-truth answer represented as a token sequence $Y = \{y_1, y_2, \ldots, y_T\}$. The model is trained to maximize the conditional likelihood of generating the correct output sequence. The objective function employed for optimization is a token-level cross-entropy loss:

\begin{equation}
\label{eq:gen_loss}
\mathcal{L}_{\text{gen}} = -\frac{1}{T} \sum_{t=1}^{T} \log P(y_t \mid X, y_{<t}),
\end{equation}
where $T$ is the target sequence length, and $P(y_t \mid X, y_{<t})$ denotes the probability of predicting token $y_t$ conditioned on the input and previously generated tokens. This loss encourages the decoder to produce fluent, contextually grounded answers consistent with the multimodal evidence.

By leveraging the pretrained VLMT backbone and adapting it to this generative question answering setting, the model effectively synthesizes information across different modalities and generates high-quality responses tailored to complex multimodal multi-hop queries.

\section{Experiments}
\label{sec:experiments}

This section presents a comprehensive evaluation of the proposed VLMT framework. The experiments are designed to assess the effectiveness of VLMT in addressing the challenges of MMQA, focusing on its ability to retrieve relevant context and generate accurate, coherent answers. Evaluations are conducted on two widely used benchmark datasets—MultimodalQA and WebQA—encompassing both retrieval and answer generation tasks. 

The experimental setup includes rigorous implementation details covering model configurations, pretraining and fine-tuning strategies, and inference settings. Ablation studies are also performed to quantify the contribution of each stage in the proposed pretraining framework. Furthermore, comparative analyses against existing state-of-the-art methods demonstrate the advantages of the proposed VLMT in both architectural design and multimodal reasoning capabilities.

\subsection{Datasets}

To evaluate the effectiveness of the proposed framework, experiments are conducted using two widely adopted benchmark datasets for multimodal multi-hop question answering: MultimodalQA and WebQA.

\subsubsection{MultimodalQA~\cite{talmor2021mmqa}}

The MultimodalQA dataset consists of multimodal question-answer pairs that require complex reasoning across diverse modalities, including text, tables, and images. It encompasses a total of 16 question types, 13 of which are compositional in nature and necessitate cross-modal retrieval and multi-hop reasoning. The dataset includes 23.8K samples for training and 2.4K samples for validation, along with relevant supporting documents and distractors. 

As the ground-truth answers for the test set are not publicly available, model performance is evaluated on the validation set, which is consistent with prior studies. Answers in MultimodalQA are typically concise, comprising single words or short phrases. The metrics used for performance evaluation are Exact Match (EM) and average F1 score.

\subsubsection{WebQA~\cite{chang2022webqa}}

The WebQA dataset includes both textual and visual information sources, in which each question is designed to be interpretable using one modality—either text or image. Approximately 44\% of the image-based queries and 99\% of the text-based queries involve multi-hop reasoning, requiring the integration of information from two or more sources. Each question is accompanied by both relevant and distractor documents across modalities, reinforcing the importance of accurate retrieval in addition to answer generation. The dataset provides 34.2K training, 5K validation, and 7.5K testing question-answer pairs.  

Answers in WebQA are complete, natural language sentences. Evaluation is conducted using two metrics: Retr-F1 and QA. Retr-F1 measures retrieval accuracy based on the overlap between retrieved and supporting sources. The QA metric combines two components: QA-FL, computed using BARTScore~\cite{yuan2021bartscore} to assess fluency and coherence between generated and reference answers, and QA-Acc, which evaluates the overlap of key entities to assess factual accuracy.

\subsection{Implementation Details}

\subsubsection{VLMT Backbone}

This study introduces two configurations of the proposed VLMT backbone, denoted as VLMT-Base and VLMT-Large. Both variants share a consistent architectural design, comprising a vision encoder and a language model. They differ primarily in model capacity, including hidden size, number of layers, and attention heads. To enable direct multimodal fusion via token-level injection, the vision encoder and language model in each variant are configured to operate with identical hidden dimensions—768 for VLMT-Base and 1024 for VLMT-Large. This design eliminates the need for projection layers, simplifying the fusion of visual and textual representations. A summary of these architectural specifications is provided in Table~\ref{tab:vlmt_configurations}.

\begin{table}[htbp]
    \centering
    \caption{VLMT model components and configurations.}
    \label{tab:vlmt_configurations}
    \renewcommand{\arraystretch}{1.5}
    \setlength{\tabcolsep}{4pt}
    \begin{tabular}{|c|c|c|c|c|c|} \hline
        \textbf{Model} & \textbf{Component} & \textbf{Hidden Size} & \textbf{\#Enc} & \textbf{\#Dec} & \textbf{\#Heads} \\ \hline
        \multirow{2}{*}{VLMT-Base}  & Vision Encoder & 768  & 12 & 0  & 12 \\ \cline{2-6}
                                    & Language Model & 768  & 12 & 12 & 12 \\ \hline
        \multirow{2}{*}{VLMT-Large} & Vision Encoder & 1024 & 24 & 0  & 16 \\ \cline{2-6}
                                    & Language Model & 1024 & 24 & 24 & 16 \\ \hline
    \end{tabular}
\end{table}

To optimize the model for downstream multimodal tasks, both variants are pretrained using the proposed three-stage framework outlined in Section~\ref{sec:vlmt_pretraining_framework}. The pretraining hyperparameters applied are consistent across both VLMT-Base and VLMT-Large, provided in Table~\ref{tab:vlmt_pretraining_hyperparameters}.

\begin{table}[htbp]
    \centering
    \caption{Pretraining hyperparameters for VLMT across three stages.}
    \label{tab:vlmt_pretraining_hyperparameters}
    \renewcommand{\arraystretch}{1.2}
    \setlength{\tabcolsep}{12pt}
    \begin{tabular}{lccc} \hline \noalign{\smallskip}
        \textbf{Hyperparameter}   & \textbf{Stage 1} & \textbf{Stage 2} & \textbf{Stage 3} \\ \noalign{\smallskip} \hline \noalign{\smallskip}
        Trainable weights         & VE               & VE \& LM         & LM               \\
        Global batch size         & 256              & 128              & 128              \\
        Epoch                     & 1                & 1                & 1                \\
        LM learning rate          & N/A              & 1e-4             & 1e-4             \\
        VE learning rate          & 1e-3             & 5e-4             & N/A              \\
        VE LLRD factor            & 0.5              & 0.5              & N/A              \\
        Weight decay              & 0.05             & 0.05             & 0.05             \\
        Scheduler                 & Cosine           & Cosine           & Cosine           \\
        Optimizer                 & AdamW            & AdamW            & AdamW            \\ \noalign{\smallskip} \hline
    \end{tabular}
    
    \vspace{0.5em}
    \begin{minipage}{0.95\linewidth}
        \small
        \textit{Note:} VE = vision encoder, LM = language model, and \textit{LLRD} refers to layer-wise learning rate decay.
    \end{minipage}
\end{table}

Following prior work~\cite{bao2021beit, dong2022clip}, layer-wise learning rate decay is applied to the vision encoder during the first two pretraining stages. This strategy assigns higher learning rate to the top layer and lower learning rate to the bottom layers, preserving low-level visual features while encouraging high-level alignment with the language model. Such differential training approach facilitates stable convergence and improved cross-modal representation learning.

\subsubsection{VLMT Reranker and Question Answering Model}

The reranker and question answering components are instantiated from the pretrained VLMT backbone, as detailed in Section~\ref{sec:vlmt_reranker} and Section~\ref{sec:vlmt_qa_model}. Each model retains the backbone’s multimodal architecture while incorporating task-specific adaptations their respective tasks.

During fine-tuning, the vision encoder is kept frozen for both models to preserve the visual representations acquired during pretraining. This design allows the optimization process to focus exclusively on the language model, ensuring stable convergence while reducing training complexity.

To promote generalization and prevent overfitting, dropout regularization is applied according to model scale. A dropout rate of 0.05 is used for VLMT-Base, while VLMT-Large adopts a higher rate of 0.1, reflecting its increased parameter capacity. This regularization strategy supports robust performance across diverse datasets and task conditions.

The fine-tuning procedure for both the reranker and QA model is standardized across the MultimodalQA and WebQA benchmarks. The key hyperparameter configurations are provided in Table~\ref{tab:mmqa_hyperparameters}, including batch sizes, learning rates, and optimization settings tailored to the distinct demands of context retrieval and answer generation.

\begin{table}[htbp]
    \centering
    \caption{Fine-tuning hyperparameters for the VLMT reranker and question answering model.}
    \label{tab:mmqa_hyperparameters}
    \renewcommand{\arraystretch}{1.2}
    \setlength{\tabcolsep}{12pt}
    \begin{tabular}{lcc} \hline \noalign{\smallskip}
        \textbf{Hyperparameter} & \textbf{Reranker} & \textbf{QA Model} \\ \noalign{\smallskip} \hline \noalign{\smallskip}
        Global batch size       & 256               & 16                \\
        Epochs                  & 3                 & 5                 \\
        Learning rate           & 2e-4              & 5e-5              \\
        Scheduler               & Cosine            & Cosine            \\
        Optimizer               & AdamW             & AdamW             \\ \noalign{\smallskip} \hline
    \end{tabular}
\end{table}

\subsubsection{Inference Settings}

During inference, a consistent retrieval strategy is adopted across all datasets to ensure comparability and fairness in evaluation. For context selection, a relative threshold of 0.5 is employed, whereby a candidate document is retained if its normalized relevance score is at least 50\% of that of the highest-scoring candidate. This dynamic thresholding mechanism allows flexibility across different question-document distributions. In conjunction with the threshold, a top-$k$ constraint is imposed with $k=5$, ensuring that only the five most relevant documents are selected for downstream processing. This dual criterion balances retrieval precision and contextual coverage, and is uniformly applied to both the MultimodalQA and WebQA datasets.

For answer generation, greedy decoding is used as the inference strategy. At each generation step, the token with the highest predicted probability is selected, yielding a deterministic and computationally efficient decoding process. The maximum generation length is set to 64 tokens for MultimodalQA and 128 tokens for WebQA, accommodating the linguistic characteristics and expected answer lengths in the respective datasets.

\subsection{Ablation Study}

To evaluate the effectiveness of the proposed three-stage pretraining strategy, a series of ablation experiments were conducted to quantify the incremental impact of each stage on multimodal retrieval and question answering performance. The results, summarized in Table~\ref{tab:ablation}, provide empirical evidence of the necessity and effectiveness of progressive multimodal alignment and task-specific adaptation.

\begin{table}[htbp]
    \centering
    \caption{Ablation results of the VLMT pretraining framework.}
    \label{tab:ablation}
    \renewcommand{\arraystretch}{1.2}
    \setlength{\tabcolsep}{6pt}
    \begin{tabular}{lccccc} \hline \noalign{\smallskip}
        \multirow{2}{*}{\textbf{Model}} & \multicolumn{3}{c}{\textbf{MultimodalQA}} & \multicolumn{2}{c}{\textbf{WebQA}} \\
        & \textbf{Retr-F1} & \textbf{EM} & \textbf{F1} & \textbf{Retr-F1} & \textbf{QA} \\ \noalign{\smallskip} \hline \noalign{\smallskip}
        VLMT-Baseline    & 82.9 & 57.6 & 62.0 & 85.7 & 39.4 \\
        + Stage 1 PT     & 88.9 & 66.7 & 70.8 & 86.5 & 41.3 \\
        + Stage 2 PT     & 89.2 & 67.9 & 71.8 & 86.6 & 42.4 \\
        + Stage 3 PT     & 89.4 & 68.9 & 72.8 & 86.9 & 42.8 \\ \noalign{\smallskip} \hline
    \end{tabular}
\end{table}

The baseline configuration (denoted as VLMT-Baseline) is initialized from independently pretrained vision and language models without any joint multimodal pretraining. This setting results in notably lower performance across both MultimodalQA and WebQA tasks, highlighting the inadequacy of relying solely on pretrained unimodal components in the absence of dedicated cross-modal alignment.

The first pretraining stage (Stage 1 PT) focuses on aligning visual embeddings from the vision encoder with the frozen embedding space of the language model. This alignment is achieved through an instruction-following paradigm based on image-caption pairs. The gains in both retrieval accuracy and question answering performance indicate that the one-sided alignment process substantially improves the model’s capacity to bridge visual and textual modalities.

The second stage (Stage 2 PT) performs joint optimization of both the vision encoder and the language model using semantically rich image-text pairs. This process strengthens fine-grained visual-semantic correspondence, allowing the model to better capture spatial, contextual, and attribute-level cues within multimodal inputs. The observed improvements in retrieval and generation metrics at this stage affirm the value of bidirectional vision-language training.

The final stage (Stage 3 PT) targets visual question answering by fine-tuning only the language model on QA supervision while keeping the vision encoder frozen. This phase enhances the model’s reasoning and answer generation capabilities without disrupting the already established visual representations. The incremental gains here are most pronounced in QA performance, demonstrating the utility of task-specific adaptation after multimodal pretraining.

Overall, the consistent performance improvements across all stages validate the proposed three-stage framework. Each stage contributes distinct and complementary benefits, with earlier stages establishing robust multimodal representations and the final stage refining task-specific reasoning.

\subsection{Experimental Results}

The effectiveness of the proposed VLMT framework is evaluated on two widely adopted MMQA benchmarks: MultimodalQA~\cite{talmor2021mmqa} and WebQA~\cite{chang2022webqa}. The evaluation focuses on both retrieval and answer generation performance, highlighting the contributions of the proposed multimodal architecture, pretraining strategy, and context selection mechanisms.

\subsubsection{Results on MultimodalQA}

Table~\ref{tab:multimodalqa_results} reports results on the validation split of the MultimodalQA dataset. VLMT-Base achieves state-of-the-art performance across all evaluation categories—single-modal, multi-modal, and overall—outperforming strong baselines including Solar~\cite{yu2023unified}, PERQA~\cite{yang2023progressive}, and UniRaG~\cite{lim2024unirag}. This improvement demonstrates the effectiveness of VLMT’s architecture, particularly the token-level injection mechanism that enables seamless fusion of visual and textual representations.

\begin{table}[htbp]
    \centering
    \caption{Experimental results on the MultimodalQA dataset (validation set).}
    \label{tab:multimodalqa_results}
    \renewcommand{\arraystretch}{1.2}
    \setlength{\tabcolsep}{6pt}
    \begin{tabular}{lcccccc} \hline \noalign{\smallskip}
        \multirow{2}{*}{\textbf{Model}} & \multicolumn{2}{c}{\textbf{Single-Modal}} & \multicolumn{2}{c}{\textbf{Multi-Modal}} & \multicolumn{2}{c}{\textbf{All}} \\
        & \textbf{EM} & \textbf{F1} & \textbf{EM }& \textbf{F1} & \textbf{EM} & \textbf{F1} \\ \noalign{\smallskip} \hline \noalign{\smallskip}
        AutoRouting~\cite{talmor2021mmqa}    & 51.7 & 58.5 & 34.2 & 40.2 & 44.7 & 51.1 \\
        ImplicitDecomp~\cite{talmor2021mmqa} & 51.6 & 58.4 & 44.6 & 51.2 & 48.8 & 55.5 \\
        SKURG~\cite{yang2023enhancing}       & 66.1 & 69.7 & 52.5 & 57.2 & 59.8 & 64.0 \\
        Solar~\cite{yu2023unified}           & 69.7 & 74.8 & 55.5 & 65.4 & 59.8 & 66.1 \\
        PERQA~\cite{yang2023progressive}     & 69.7 & 74.1 & 54.7 & 60.3 & 62.8 & 67.8 \\
        UniRaG~\cite{lim2024unirag}          & 71.7 & 75.9 & 62.3 & 66.0 & 67.4 & 71.3 \\ \noalign{\smallskip} \hline \noalign{\smallskip}
        VLMT-Base                             & 74.2 & 78.1 & 62.6 & 66.6 & 68.9 & 72.8 \\
        VLMT-Large                            & 80.5 & 84.4 & 71.9 & 75.0 & 76.5 & 80.1 \\ \noalign{\smallskip} \hline
    \end{tabular}
\end{table}

VLMT-Large further improves performance by leveraging increased model capacity and richer representation depth. With its larger hidden size and more attention heads, VLMT-Large achieves notable improvements in EM and F1, particularly in multi-modal scenarios where fine-grained alignment and reasoning are essential. The outstanding results underscore the scalability of the proposed framework and validate the design choice of consistent hidden dimensions across vision encoder and language model components.

\subsubsection{Results on WebQA}

Table~\ref{tab:webqa_results} presents results on the WebQA test set. The dataset poses additional challenges due to its open-domain nature and the requirement to generate fluent, full-sentence answers. VLMT-Base outperforms other base-sized models such as MuRAG~\cite{chen2022murag}, SKURG~\cite{yang2023enhancing}, and Solar~\cite{yu2023unified} in terms of QA-FL and QA-Acc metrics, reflecting its ability to effectively integrate visual and textual information through the pretrained VLMT backbone.

\begin{table}[htbp]
    \centering
    \caption{Experimental results on the WebQA dataset (test set).}
    \label{tab:webqa_results}
    \renewcommand{\arraystretch}{1.2}
    \setlength{\tabcolsep}{6pt}
    \begin{tabular}{lcccc} \hline \noalign{\smallskip}
        \textbf{Model} & \textbf{Retr-F1} & \textbf{QA-FL} & \textbf{QA-Acc} & \textbf{QA} \\ \noalign{\smallskip} \hline \noalign{\smallskip}
        VLP~\cite{chang2022webqa}           & 68.9 & 42.6 & 36.7 & 22.6 \\
        VLP + VinVL~\cite{chang2022webqa}   & 70.9 & 44.2 & 38.9 & 24.1 \\
        MuRAG~\cite{chen2022murag}          & 74.6 & 55.7 & 54.6 & 36.1 \\
        SKURG~\cite{yang2023enhancing}      & 88.2 & 55.4 & 57.1 & 37.7 \\
        Solar~\cite{yu2023unified}          & 89.4 & 60.9 & 58.9 & 40.9 \\
        PERQA~\cite{yang2023progressive}    & 89.6 & 61.7 & 63.9 & 44.4 \\ \noalign{\smallskip} \hline \noalign{\smallskip}
        VLMT-Base                            & 86.9 & 61.1 & 61.4 & 42.8 \\
        VLMT-Large                           & 87.8 & 64.0 & 66.7 & 47.6 \\ \noalign{\smallskip} \hline
    \end{tabular}
\end{table}

Nevertheless, VLMT-Base shows slightly lower performance in QA compared to VLMT-Large, owing to limitations in generative capacity. The improvements achieved by VLMT-Large demonstrate the efficacy of scaling the architecture, which not only enhances fluency and factual accuracy but also strengthens multimodal reasoning.

While the retrieval metric (Retr-F1) of VLMT remains slightly lower than some baselines, this is a result of the proposed relative threshold and top-$k$ retrieval strategy, which prioritizes retrieval recall. This approach increases the diversity and completeness of the retrieved evidence, providing broader contextual grounding for the QA model at the cost of retrieval precision. However, the richer context positively contributes to downstream answer generation, as evidenced by the superior QA scores of VLMT-Large.

\section{Conclusion}
\label{sec5:conclusion}

This paper presents Vision-Language Multimodal Transformer (VLMT), a unified and scalable framework designed to address the challenges of multimodal multi-hop question answering (MMQA). VLMT integrates a transformer-based vision encoder with a sequence-to-sequence language model within a shared embedding space, enabling direct token-level fusion of visual and textual representations. A three-stage pretraining framework is introduced to progressively align vision-language representations and enhance reasoning capabilities, significantly improving the model’s ability to process and synthesize multimodal evidence.

Built on the pretrained VLMT backbone, a two-stage framework with task-specific modules are developed: a multimodal reranker and a multimodal question answering (QA) model. The reranker predicts document relevance scores and utilizes a relative threshold with top-$k$ selection strategy, ensuring the retrieval of diverse and informative contexts. The QA model performs answer generation by attending to both the input question and retrieved multimodal content.

Extensive experiments on two benchmark datasets, MultimodalQA and WebQA, prove the effectiveness of the proposed approach. On MultimodalQA, VLMT-Large achieves 80.5 EM and 84.4 F1 in single-modal settings, and 71.9 EM and 75.0 F1 in multi-modal settings, outperforming prior state-of-the-art methods including UniRaG and PERQA. Similarly, on WebQA, VLMT-Large achieves 64.0 QA-FL and 66.7 QA-Acc, surpassing the best-performing baseline, PERQA, and setting a new benchmark for multimodal answer generation. These results underscore the advantages of the proposed architecture, pretraining methodology, and retrieval mechanism.

Looking forward, several directions remain open for future research. First, enhancing retrieval precision while maintaining high recall remains a key challenge in this domain. More adaptive retrieval strategies that dynamically balance relevance and diversity could further improve the overall QA performance. Second, extending VLMT to handle additional modalities such as audio or video may broaden its applicability in real-world multimodal information systems. Finally, investigating the implementation of lightweight or distilled variants of VLMT could facilitate deployment in resource-constrained environments while preserving performance.



\bibliographystyle{ieeetr}
\bibliography{references}

\begin{IEEEbiography}[{\includegraphics[width=1in,height=1.25in,clip,keepaspectratio]{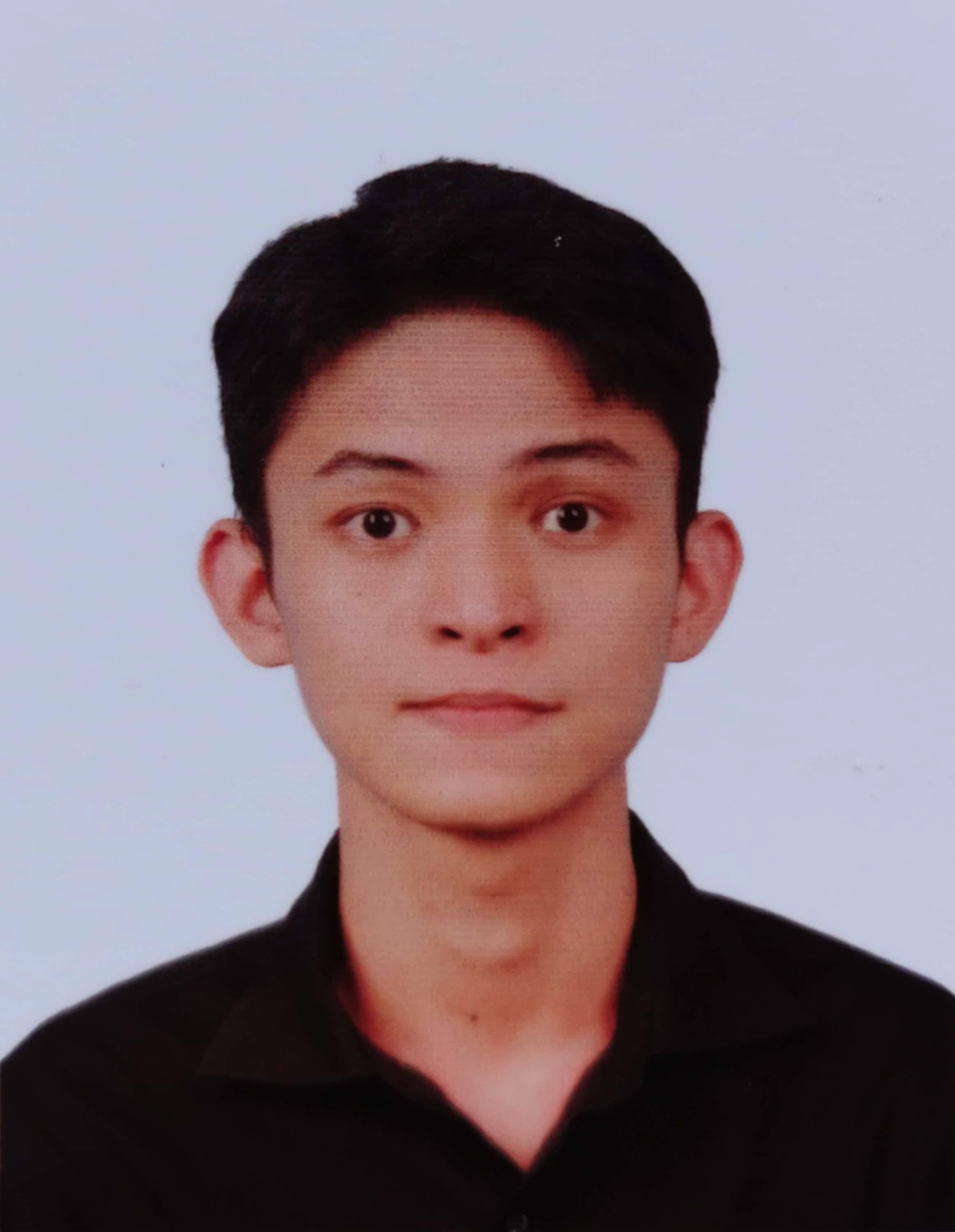}}]{Qi Zhi Lim}
received his Bachelor’s degree in Computer Science (Hons.) Artificial Intelligence from Multimedia University, Malaysia, in 2023. He is currently pursuing a Ph.D. degree in Information Technology, focusing on multimodal multi-hop question answering. His research interests include multimodal data processing, feature extraction and integration, information retrieval, and question answering.
\end{IEEEbiography}

\begin{IEEEbiography}[{\includegraphics[width=1in,height=1.25in,clip]{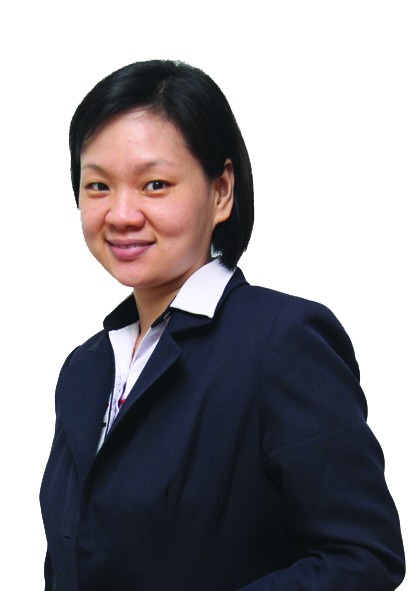}}]{Chin Poo Lee} (Senior Member, IEEE)
received the M.Sc. and Ph.D. degrees in Information Technology, specializing in abnormal behavior detection and gait recognition. She is currently an Assistant Professor with the School of Computer Science, University of Nottingham Ningbo China. Her research interests include computer vision, natural language processing, and deep learning.
\end{IEEEbiography}

\begin{IEEEbiography}[{\includegraphics[width=1in,height=1.25in,clip,keepaspectratio]{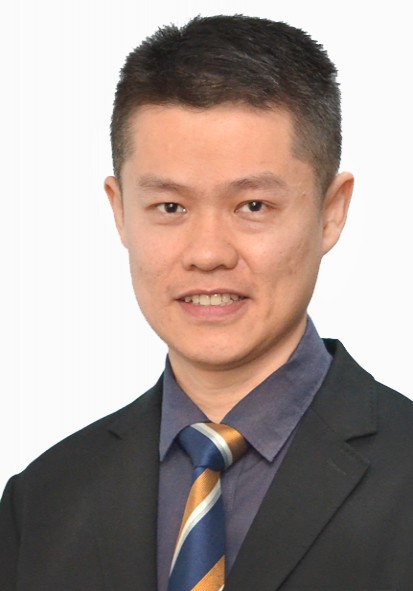}}]{Kian Ming Lim} (Senior Member, IEEE) received the B.IT. (Hons.) degree in Information Systems Engineering, the Master of Engineering Science (M.Eng.Sc.) degree, and the Ph.D. degree in Information Technology from Multimedia University, Malaysia. He is currently an Associate Professor with the School of Computer Science, University of Nottingham Ningbo China. His research interests include machine learning, deep learning, computer vision, and pattern recognition.
\end{IEEEbiography}

\begin{IEEEbiography}[{\includegraphics[width=1in,height=1.25in,clip,keepaspectratio]{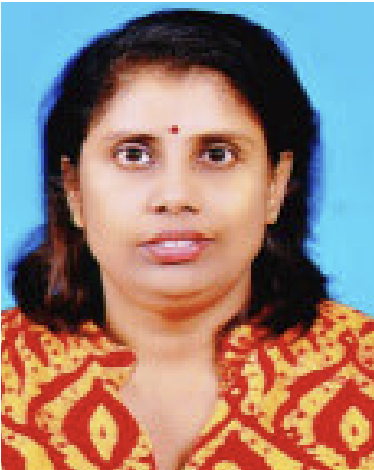}}]{Kalaiarasi Sonai Muthu Anbananthen} received the Ph.D. degree in Artificial Intelligence. She is currently a Professor with the Faculty of Information Science and Technology, Multimedia University, Malaysia. She has served as the Program Coordinator for the Master of Information Technology (Information Systems) and as the Co-coordinator for the Business Intelligence and Analytics (BIA) program. She is a reviewer for various Scopus- and SCI-indexed technical journals. She has secured multiple national and international research grants as Principal Investigator and has published more than 120 journal articles, conference papers, and book chapters. Her current research interests include data mining, sentiment analysis, artificial intelligence, machine learning, deep learning, and text analytics.
\end{IEEEbiography}

\end{document}